# THINKING WITH MANY MINDS:
# USING LARGE LANGUAGE MODELS
# FOR MULTI-PERSPECTIVE PROBLEM-SOLVING

Sanghyun Park (National University of Singapore)
shpark@nus.edu.sg

Boris Maciejovsky (UC Riverside)
borism@ucr.edu

Phanish Puranam (INSEAD)
phanish.puranam@insead.edu

## ABSTRACT

We argue that solving complex problems involving multiple attributes and multiple actors demands cognitive flexibility—the capacity to maintain and synthesize multiple, often conflicting perspectives—yet such flexibility is difficult to achieve and sustain within a single mind. We introduce synthetic deliberation, an approach that leverages large language models (LLMs) to simulate dialogue between synthetic agents, each embodying a distinct perspective. By externalizing these perspectives and making their interactions observable, synthetic deliberation enables decision makers to preserve viewpoint diversity, control the timing and extent of integration, and explore a broader solution space than unaided reasoning typically allows. Our theoretical framework identifies the mechanisms—externalizability and tunability—through which synthetic deliberation can outperform imagined deliberation and specifies the problem conditions under which these advantages are greatest. We also examine potential downsides, including over-reliance, cognitive offloading, and ethical risks, and outline design principles to ensure the technology enhances rather than erodes human cognitive capability.

# 1. INTRODUCTION

Many of society's most critical challenges—whether in policy, strategy, or organizational design—involve both multiple attributes, with tradeoffs across diverse criteria (Keeney & Raiffa, 1993) and multiple stakeholders, with divergent objectives (Rittel & Webber, 1973). For decision makers, contributing to solving such problems demands *cognitive flexibility* - the ability to transition seamlessly between diverse concepts or viewpoints (Diamond, 2013) as one imagines possible solutions. It enables problem-solvers to benefit from imagining multiple possibilities, such as toggling between abstract vs. concrete, structural vs. functional, their own vs. others' perspectives, or employing approaches like goal-driven or data-driven methods (Krems, 2014). Cognitive flexibility creates the advantages of parallelism and diversity of perspectives that underpin the "wisdom of crowds" (Page, 2000) within a single individual, allowing the emergence of "inner-crowd wisdom" (Herzog & Hertwig, 2014). However, cognitive flexibility is both difficult to achieve and hard to maintain. As F. Scott Fitzgerald (1936) famously noted, "The test of a first-rate intelligence is the ability to hold two opposed ideas in mind at the same time and still retain the ability to function."

In this paper, we make two key contributions regarding cognitive flexibility and its enhancement. First, we introduce a novel dual-process framework explicitly theorizing cognitive flexibility as emerging at the aggregate level from the interplay between two complementary sub-processes—*compartmentalization* and *integration*. Compartmentalization segments conflicting perspectives (Jonassen, 2000; Newell & Simon, 1972), while integration reconciles them through dynamic reasoning and problem synthesis (Holyoak & Thagard, 1997; Guilford, 1967; Johnson-Laird, 2010). Using a formal framework of multiagent collaborative search on rugged landscapes, we explain why cognitive flexibility at the level of the system of agents can emerge even if no individual agent has it. However, humans typically struggle to compartmentalize and integrate differing perspectives iteratively (for instance, by imagining different stakeholders engaging in deliberation), though this capability is variable among individuals and can be cultivated to some extent through practice (Showers, 1992).

Second, we describe **synthetic deliberation** as a promising approach to overcome the challenges in achieving cognitive flexibility. We define synthetic deliberation *as a technology-supported process (specifically using Large Language Models, henceforth LLMs) that simulates discourse between synthetic agents representing diverse perspectives on a problem*. We theorize how synthetic deliberation can outperform imagined deliberation through two key mechanisms—externalizability and tunability—and specify the boundary conditions under which these advantages are most pronounced. While synthetic deliberation shares methodological roots with simulation, digital twinning, and agent-based modeling, it is distinct in its structure and purpose. Rather than predicting outcomes (as in simulation) or establishing a direct correspondence between virtual and real-world systems (as in digital twinning) (Lyytinen, Weber, Becker, & Pentland, 2023), synthetic deliberation aims to simulate deliberative dialogue to enhance human cognitive flexibility. It also differs from agent-based modeling (Knudsen, Levinthal, & Puranam, 2019), where the objective is to explain the emergence of complexity from the aggregation of agents following relatively simple rules. Instead, synthetic deliberation aims to provide a platform for maintaining and integrating divergent viewpoints in complex problems. It is thus an approach to improve an individual's capacity to effectively "think with many minds."

Our concept of synthetic deliberation aligns with recent proposals to use multiple LLMs to surface disparate views, clarify objectives (Burton et al., 2024), and improve human deliberative quality through rephrasing (Argyle, Busby, Gubler, Bail, Howe, Rytting, & Wingate, 2023). We also build on recent evidence showing that LLMs can reasonably replicate individuals' attitudes and behaviors, offering confidence in their ability to simulate nuanced perspectives and attitudes (Park et al., 2024; Anthis et al., 2025; Kozlowski & Evans, 2025). We extend this line of work by offering a theoretical framework for how simulated deliberation can enhance individual cognitive flexibility through externalization (to overcome cognitive limitations in reserving compartmentalization) and tunable control over integration. Tunability is analogous, in single-agent settings, to varying the weights assigned to different objectives in multi-criteria reasoning (Keeney & Raiffa, 1993), and, in multiagent settings, to experimenting with different levels of willingness to compromise and reach agreement among stakeholders with divergent objectives (Rittel & Webber, 1973).

We believe that our approach offers substantial potential for enhancing decision-making in practice. In strategic planning, managers can simulate the interplay of diverse viewpoints to refine strategic options and anticipate tradeoffs. Policymakers can assess the effects of proposed measures across heterogeneous stakeholder groups. In conflict resolution, synthetic deliberation can model negotiations to identify mutually beneficial compromises. In each case, the goal is the same: to support humans in sustaining, exploring, and synthesizing multiple perspectives on complex problems more effectively than they could alone.

The remainder of this paper proceeds as follows. In Section 2, we establish the theoretical background of the concept of cognitive flexibility, emphasizing the challenges in simultaneously achieving compartmentalization and integration within a single mind. Section 3 introduces a formal model of multiagent deliberation, providing the foundation for understanding how cognitive flexibility can emerge at the group level through iterative interactions between agents holding diverse perspectives. Building on this model, Section 4 presents our propositions about the conditions under which synthetic deliberation enhances individual problem-solving capability by focusing on problem complexity, tunability, and learning. Finally, Section 5 discusses theoretical and practical implications, compares synthetic deliberation with related approaches, and addresses limitations and future directions.

## 2. THEORETICAL BACKGROUND

### 2.1. Cognitive Flexibility in Solving Complex Problems: Challenges

Problem-solving fundamentally relies on the capacity to mentally simulate different courses of action to assess their likelihood of success (Szpunar, Spreng, & Schacter, 2014). By constructing a mental representation of a situation and virtually experiencing it through sensory, emotional, and cognitive dimensions, mental simulations allow individuals to anticipate outcomes, strategize, and prepare for various possibilities (Galinsky, Wang, & Ku, 2005; Parker, Atkins, & Axtell, 2008). This process can be likened to individuals navigating a fitness landscape in their minds to identify promising points (Thagard, 2005).

The process of mental simulation faces unique challenges when applied to complex problems. These problems can be visualized as "rugged fitness landscapes," characterized by intricate interdependencies among components (i.e., complexity) that generate numerous local peaks. Navigating these landscapes to identify globally optimal solutions is notably challenging, as their complexity frequently traps decision-makers prone to local search in suboptimal outcomes (Rittel & Webber, 1973; Levinthal, 1997). When complex problems involve diverse stakeholders with conflicting viewpoints, exploring such landscapes may become even more intricate because the structure of interdependencies may produce uncorrelated (or even misaligned) fitness landscapes across actors (Rivkin & Siggelkow, 2003; Koçak, Levinthal, & Puranam, 2023b). Navigating such complexity necessitates cognitive mechanisms that enable problem-solvers to break free from entrapment on local peaks and explore distant solution spaces.

In principle, compartmentalization can allow individuals to mentally isolate conflicting perspectives, minimizing interference and enabling parallel exploration without one set of assumptions contaminating another. This can help expand the scope of the search for possible solutions. The logic of compartmentalization is also related to the "devil's advocate" approach in problem-solving. Research demonstrates that exposure to opposing viewpoints enhances information-seeking, strategy diversity, and idea generation (Nemeth & Rogers, 1996; Nemeth & Kwan, 1987; Nemeth, 1995). Both explicit instructions and indirect strategies for considering alternative perspectives reduce biased inference (Lord, Lepper, & Preston, 1984). In legal settings, professionals routinely employ counter-arguments to enhance evaluative rigor, highlighting the value of systematically engaging with distinct viewpoints (Burke, 2006; Lidén, Gräns, & Juslin, 2019).

In contrast to compartmentalization, integration involves synthesizing divergent perspectives into a cohesive and more comprehensive understanding. It entails combining the results of two or more compartmentalized simulations (Siggelkow & Levinthal, 2003). This capacity for integration is closely linked to the concept of knowledge recombination, which has been studied extensively in the literature on organizations (e.g., Kogut & Zander, 1993; Ahuja & Katila, 2004; Kaplan & Vakili, 2015; Fleming & Sorenson, 2004; Karim & Kaul, 2015; Rosenkopf & Nerkar, 2001; Yayavaram & Ahuja, 2008). This literature shows that successful integration through knowledge recombination requires both the ability to identify valuable combinations and the capacity to effectively synthesize them into new understandings.

In summary, compartmentalization reduces interference among competing viewpoints,

enabling decision-makers to explore multiple alternatives in parallel (Siggelkow & Levinthal, 2003). Integration, by contrast, leverages these divergent explorations by synthesizing them into novel solutions that surpass the value of any single perspective alone. Notably, integration is effective only when the diversity of perspectives is preserved and not prematurely discarded. Page (2007) highlights the crucial role of maintaining diverse perspectives in fostering integrative problem-solving, often yielding outcomes superior to those achieved by individuals with objectively higher abilities (also see Hong & Page, 2001, 2004).

However, maintaining the delicate balance between the dual processes of compartmentalization and integration—both critical components of cognitive flexibility—is challenging. On the one hand, both compartmentalization and integration of different perspectives are subject to socio-cognitive constraints inherent in individual decision-making. For instance, individuals are naturally inclined to seek confirmatory evidence that reinforces their existing beliefs about judgments, predictions, or decisions. Even when these beliefs are spurious, they persist because they offer a comforting sense of causal understanding about the world. Anderson and Sechsler (1986) showed that merely explaining a potential link between variables increases individuals' confidence in their beliefs. This tendency, often rooted in confirmation bias (e.g., Nickerson, 1998), can inhibit the exploration and integration of alternative perspectives. On the other hand, while mental compartmentalization and integration functionally complement each other, their operations may interfere with each other. For instance, the integration of diverse viewpoints may result in premature convergence in parallel exploration (e.g., Park & Puranam, 2024), whereas the compartmentalization of different perspectives may hinder their timely integration.

## 2.2. Cognitive Flexibility in Solving Complex Problems: Current Solutions

Training can improve cognitive flexibility for problem-solving (Buitenweg, Van de Ven, Prinssen, Murre, & Ridderinkhof, 2017). Experimental evidence suggests that when individuals are instructed to engage in "counter-explanations," it effectively reduces cognitive bias in exploring and integrating different perspectives (Van Brussel, Timmermans, Verkoeijen, & Paas, 2020). Similar debiasing effects have been observed in addressing anchoring (Mussweiler, Strack, & Pfeiffer, 2000), overconfidence (Griffin, Dunning, & Ross, 1990), and hindsight biases (Arkes, Faust, Guilmette, & Hart, 1988). These strategies not only counteract cognitive biases but also enhance information-seeking, strategy diversity, and originality in problem-solving (Nemeth, 1995; Nemeth & Rogers, 1996).

Dialogue-based learning is particularly relevant to our argument. Chi, Kang, and Yaghmourian (2017) found that dialogue formats outperform monologues for learning complex concepts because they naturally surface competing viewpoints and keep them cognitively distinct. This structure effectively compartmentalizes perspectives, enabling parallel mental simulations without interference—a core mechanism of cognitive flexibility. Related work on vicarious learning through dialogue shows that observers can achieve learning gains comparable to direct participants (Craig, Driscoll, & Gholson, 2004; Driscoll, Craig, Gholson, Hu, & Grasessner, 2003). Observing episodes of conflict can further enhance learning (Schunk, Hanson, & Cox, 1987) by providing *refutation information* (Muller, Bewes, Sharma, & Reimann, 2008), and Chi et al. (2017) report that observers often engaged differently with conflicting perspectives than the dialogue participants themselves, suggesting that observation provides unique cognitive advantages in processing diverse viewpoints.

However, the practical application of such approaches often encounters significant challenges. The most obvious is simply the cost of arranging for others to engage in a dialogue for the benefit of an observer. An alternative may be to mentally simulate such a deliberative process: just as one takes on a devil's advocate role, one might imagine mentally simulating multiple personalities, each arguing from a different perspective. Yet, this too quickly runs up against cognitive limits: working memory capacity (Cowan, 2001) constrains how many perspectives can be maintained and switched between, especially under time pressure. Complex or unfamiliar problems amplify the burden, and simulating multiple actors often yields less robust and less sustained mental models (Baddeley & Hitch, 1974).

Moreover, even when individuals successfully simulate a multi-perspective dialogue, the process remains vulnerable to biases. The drive for internal consistency (Nickerson, 1998) can suppress exploration of genuinely conflicting viewpoints, and the path-dependent nature of thought can cause one perspective to contaminate another. These factors limit the viability of purely mental approaches for sustaining compartmentalization and integration over time.

In the following sections, we introduce a formal framework to first describe how deliberation processes between multiple agents can generate cognitive flexibility at the group level. We then propose that synthetic deliberation—which leverages Artificial Intelligence (AI) to create a discussion among artificial agents that each represents a perspective on a problem—provides a powerful tool to improve cognitive flexibility in solving complex problems.

## 3. A DUAL-PROCESS MODEL OF COGNITIVE FLEXIBILITY BASED ON MULTIAGENT DELIBERATION

To rigorously theorize about the dual processes of compartmentalization and integration that produce cognitive flexibility, we formalize these within the framework of multiple agents searching in parallel and integrating their efforts (e.g., Siggelkow & Levinthal, 2003; Koçak,

Levinthal, & Puranam, 2023a; Park & Puranam, 2024). Specifically, we build upon the framework introduced by Holland (1975) and Levinthal (1997), in which agents with different perspectives collectively navigate the solution space of complex problems. Using this framework, we compare unaided mental simulation with synthetic deliberation by explicitly representing both the multiplicity of perspectives and their integration over time.

The model consists of three components: the problem space, agents with diverse perspectives, and model dynamics. In Appendix A, we show simulation results from the computational implementation of our model.

### 3.1. Problem Spaces

We represent the problem space as a landscape $\mathcal{L}$ with an objective payoff function $\Pi(\mathbf{x})$ for any position $\mathbf{x} \in \mathcal{L}$, which can be a multi-dimensional vector. Here, $\Pi(\mathbf{x})$ indicates the aggregate-level fitness (or welfare), serving as the decision-maker (DM)'s ground truth payoff. Following prior work (e.g., Kauffman, 1993; Levinthal, 1997), we assume that payoffs for proximate solutions are not necessarily correlated. Consequently, the fitness landscape may feature multiple peaks (instead of a single peak) due to interdependence and tradeoffs between attributes. In other words, no additive separability of utility is assumed.

### 3.2. Agents with Diverse Perspectives

We assume the existence of $m \in \mathbb{N}$ agents, indexed by $i \in \{1, ..., m\}$, initially randomly assigned to positions (i.e., $\mathbf{x}_i$ for $i \in \{1, 2, ..., m\}$) on the landscape $\mathcal{L}$. In addition, we allow agents to hold heterogeneous beliefs on possible solutions, which may arise from having either diverging value systems or a limited understanding of complex environments. We thus represent an individual agent's belief as $\pi_i(\mathbf{x})$ for $i \in \{1, 2, ..., m\}$—which may deviate not only from the aggregate-level payoff $\Pi(\mathbf{x})$ but also from others' beliefs (i.e., $\pi_i(\mathbf{x}) \neq \pi_j(\mathbf{x})$ for $j \in \{1, 2, ..., m\} - \{i\}$) due to their distinctive views. We assume that the agent $i$'s perceived payoff can be formally represented as:

$$\pi_i(\mathbf{x}) = f(\Pi(\mathbf{x}), \beta_i(\mathbf{x})), \tag{1}$$

where $\beta_i(\mathbf{x})$ represents a systematic divergence in agents' evaluations of the alternatives. When navigating the fitness landscape, individual agents in our model act upon an incomplete representation of the solution space in that their consideration sets, $\ell(\mathbf{x})$, are restricted (i.e., $|\ell(\mathbf{x})| < |\mathcal{L}|$). Specifically, we assume that individual agents only consider proximate alternatives (i.e., local search). Formally, the consideration set of an agent $i$ at period $t$ is represented as:

$$\ell(\mathbf{x}_{it}) = \{\mathbf{x} | D(\mathbf{x}, \mathbf{x}_{it}) \leq d\}, \tag{2}$$

where $D(\mathbf{x}, \mathbf{x}_{it})$ is the distance between $\mathbf{x}$ and $\mathbf{x}_{it}$, and $d$ is a parameter that tunes the search scope. This consideration set is an agent's perspective.

### 3.3. Model Dynamics

We assume that agents search over $T$ discrete rounds, indexed by $t \in \{1, \ldots, T\}$. Each round consists of two phases corresponding to the dual processes in our theory.

<u>Phase 1: Compartmentalized local search</u>

First, individual agents conduct local search independently until they reach their respective local optima, such that there is no better-off solution in their consideration sets (i.e., parallel search). Formally, the process of local search for an agent $i$ can be represented as:

$$\mathbf{x}_{it(z+1)} = \underset{\mathbf{x}}{\mathrm{argmax}}\{\pi_i(\mathbf{x}) | \mathbf{x} \in \ell(\mathbf{x}_{itz})\}, \tag{3}$$

where $z$ indicates the number of local search steps taken ($\mathbf{x}_{it0} = \mathbf{x}_{it}$). This process repeats until the agent $i$ reaches a local peak (i.e., $\mathbf{x}_{it(z'+1)} = \mathbf{x}_{itz'}$), which becomes its pre-integration position at $t+1$. This represents the compartmentalized search by each agent for peaks within their perspectives and prevents contamination of viewpoints.

Phase 2: Integration

The next phase represents integration. In this phase, one agent $j$ is randomly selected to propose a solution based on its current pre-integration position, $\mathbf{x}_{j(t+1)}$, and the other agents integrate the proposed solution with their current solutions. This models how humans mentally simulate others' evaluations and reactions to the proposal—a process of perspective-taking. In the model, agents move toward the proposed solution at a rate of $\alpha_{j \to i}$.[1] Formally, we represent their movements as:

$$\mathbf{x}'_{i(t+1)} = (1 - \alpha_{j \to i})\mathbf{x}_{i(t+1)} + \alpha_{j \to i}\mathbf{x}_{j(t+1)} \qquad (4)$$

where $i \in \{1, \dots, m\} - \{j\}$, $\alpha \in [0, 1]$, $\mathbf{x}'_{i(t+1)}$ indicates an agent $i$'s position post-integration, and $\alpha$ is an integration parameter. Note that Equation (4) is the weighted average of the two beliefs represented as their positions, which is congruent with the DeGroot model (DeGroot, 1974) that has been used to explain empirical patterns in social influence processes (Friedkin et al., 2019; see Mastroeni et al., 2019 for a review).

This iterative process repeats until the agents reach a consensus (i.e., converge on a specific solution) or the terminal round, $T$, is reached. Then, the final decision $\mathbf{x}_{DM}$ is determined by the DM selecting the best solution (according to their own beliefs about the landscape) from among all solutions discovered throughout the process.

$$\mathbf{x}_{DM} = \underset{\mathbf{x}}{\mathrm{argmax}}\{\pi_{DM}(\mathbf{x}) | \mathbf{x} \in \{\mathbf{x}_{it} | i \in \{1, \dots m\}, t \in \{1, \dots, T\}\}\} \qquad (5)$$

In this framework, cognitive flexibility at the system level—the DM's ability to access,

---

[1] While we assume that agents are willing to accommodate others' perspectives (when $\alpha > 0$), we can modify the rule by letting agents integrate others' proposals only when they believe doing so will improve their solutions. The process of integration helps agents escape local peaks, and thus its benefit still exists regardless of whether it is motivated by self-interest or not.

preserve, and synthesize diverse perspectives—emerges from the interaction of two tunable parameters: the number of agents $m$ and the integration rate $\alpha$. The number of agents governs the extent of compartmentalization. A larger $m$ means that more distinct perspectives are maintained in parallel, increasing coverage of the problem landscape and the likelihood that at least one agent will discover a valuable region unreachable by others through local search alone (see Kauffman, 1993; Levinthal, 1997). The integration rate $\alpha$ determines how strongly agents adjust toward each other's solutions in each round.

Neither parameter is necessarily optimal to be set at its extreme values. Increasing $m$ means that each individual agent may speak and influence others infrequently, so that promising solutions may take longer to propagate through the system. An integration rate of $\alpha = 0$, meaning no movement toward others' solutions, may preserve diversity indefinitely, but it prevents any recombination of ideas (a.k.a. recombination; see Holland, 1975; March, 1991). Without integration, the DM fails to exploit complementarities between different regions of the landscape, and valuable intermediate solutions that lie "between" agents' perspectives will never be discovered. If $\alpha = 1$, all agents become homogenous in their perspective, which will also limit search. Cognitive flexibility, therefore, arises not from maximizing either parameter independently but from balancing them.

In the following section, we present propositions that specify the conditions under which finding this balance through synthetic deliberation offers advantages over attempting to do so through individual mental simulation.

## 4. THE BENEFITS OF SYNTHETIC OVER IMAGINED DELIBERATION

The preceding section described a general formalization of compartmentalization (shaped by the parameter $m$, the number of agents) and integration (shaped by the parameter $\alpha$) to jointly produce the cognitive flexibility needed to search a complex problem landscape. This process can, in principle, be mentally simulated by a human decision-maker.

We characterize "imagined deliberation" as a process in which an individual decision-maker creates $m$ imagined agents, each assigned a distinct perspective and engages in simulating deliberation between them, in their minds. This corresponds to individuals metaphorically stepping into the shoes of multiple stakeholders (e.g., CEO, local community, environmentalist, government) to generate suggestions and reasoning as well as reactions based on each stakeholder's viewpoint, followed by integrating ideas from diverse perspectives.

In contrast, in synthetic deliberation, the decision-maker acts as an observer and delegates the articulation of perspectives to AI agents such as Large Language Models (LLMs). This can be achieved either by assigning a distinct LLM to each perspective or instructing a single LLM to simulate $m$ distinct personas while maintaining compartmentalization of information. A key assumption underlying synthetic deliberation is that "an AI agent can simulate the arguments of an agent with a particular perspective *at least as well as* a human can mentally simulate such an agent." Recent research provides evidence for LLMs' capability to function as "synthetic subjects" in surveys and experiments (Horton, 2023; Mannekote et al., 2024; Tranchero et al., 2024) and as "synthetic scientists" making predictions about outcomes (Manning et al., 2024; Lippert et al., 2024; Luo et al., 2024). While LLMs may face limitations from training data that may yield biased representations (Parikh, Teeple, & Navathe, 2019), our framework only requires that LLMs match human capability to *simulate* stakeholder interactions in their minds, not that they perfectly replicate actual human behavior.

Nevertheless, as we will discuss, while the assumption of comparable accuracy in simulation is sufficient for our results, it is not strictly necessary. Even if an LLM is less effective than humans at simulating a particular perspective, synthetic deliberation can still outperform alternative approaches, such as imagined deliberation, under certain conditions that we identify.

We now articulate the merits of synthetic deliberation over imagined deliberation.

**4.1 Externalization: Preserving Perspective Diversity through Compartmentalization**

Imagined deliberation depends on the decision maker's ability to mentally simulate distinct perspectives without interference. Humans can attempt this by compartmentalizing perspectives—mentally toggling between them—but cognitive constraints on working memory (Cowan, 2001), attention, and self-distancing make it difficult to sustain such separations. As the number of perspectives grows ($m$ in our model), particularly when they are mutually incompatible or built on different premises, individuals struggle to maintain the epistemic boundaries needed to prevent premature convergence. These challenges are compounded by well-documented tendencies toward convergence bias, representational collapse, and confirmation-driven synthesis (Nickerson, 1998). Even deliberate strategies such as adopting a devil's advocate stance or constructing counter-arguments (Nemeth & Rogers, 1996; Lord, Lepper, & Preston, 1984) cannot fully prevent interference between imagined perspectives, which narrows the search space and suppresses the formation of novel combinations.

Synthetic deliberation overcomes this limitation by externalizing perspectives into $m$ distinct, role-bound agents—each with its own internally coherent evaluative function and independent search trajectory. Because each synthetic agent draws on the computational power of an underlying Large Language Model, and operates autonomously, its exploration is not contaminated by others' intermediate conclusions, avoiding the path dependencies and inadvertent cognitive blending that often occur in human-only reasoning. This architecture allows parallel searches to be maintained across different regions of the problem space, preserving diversity until the point of controlled integration.

By externalizing the representation and exploration of multiple perspectives to technologies such as Large Language Models (LLMs), synthetic deliberation enables a form of cognitive parallelism that imagined deliberation cannot easily replicate. This parallelism expands the range of candidate solutions available for later integration, increasing the potential for novel, high-quality outcomes.

**Proposition 1**: Synthetic deliberation will outperform imagined deliberation when performance depends on maintaining multiple conflicting perspectives without interference or decay during complex problem solving.

*4.2. Tunability: Dynamic Adjustment of Integration.*

Beyond preserving multiple perspectives, effective problem solving often requires knowing when and how to integrate them—a capability that imagined deliberation struggles to achieve. In imagined deliberation, decision makers cannot easily control how and when different perspectives are integrated. Once multiple viewpoints are mentally simulated, they tend to influence each other immediately, leading either to premature convergence—when integration happens too soon—or to persistent fragmentation—when integration is delayed indefinitely. This difficulty reflects broader cognitive constraints on metacognitive control, working memory, and attentional allocation (Cowan, 2001; Nickerson, 1998). Humans rarely have the bandwidth or self-regulatory capacity to adjust the balance between compartmentalization and integration dynamically during the course of reasoning.

Synthetic deliberation overcomes this limitation through tunability—in our model, for a given number of perspectives $m$, the ability to deliberately vary $\alpha$ within or across runs—allowing decision makers to control the transition from compartmentalization to integration in ways that imagined deliberation cannot. As we have noted, a low $\alpha$ preserves distinctiveness across viewpoints by minimizing mutual influence, allowing each agent to explore its conceptual subspace more thoroughly. By contrast, a high $\alpha$ accelerates convergence, producing more unified and internally coherent outcomes but at the cost of prematurely reducing diversity.

There are two forms of tuning feasible with synthetic deliberation that give it an advantage over imagined deliberation. Tuning can occur within-runs: synthetic deliberation enables dynamic modulation of $\alpha$ during the course of deliberation. For instance in a *simulated annealing* approach (Černý, 1985; Kirkpatrick, Gelatt, & Vecchi, 1983; Van Laarhoven & Aarts, 1987; for annealing in human behaviors, see Cagan & Kotovsky, 1997; Kotovsky, Hayes, & Simon 1985; Kotovsky & Simon, 1990), $\alpha$ can start low to encourage exploratory divergence and be increased later to promote integrative convergence (see Figure A1 in Appendix A). This sequencing mirrors the ideal two-stage model of deliberation—divergence followed by convergence—avoiding premature consensus while ensuring that integration occurs only after a rich set of alternatives has been explored. Fixed-$\alpha$ processes cannot achieve this temporal orchestration: high $\alpha$ from the outset collapses diversity too soon, while low $\alpha$ throughout impedes consensus formation.

Tuning can also occur across "runs". We can simulate synthetic deliberation multiple times with different fixed values of $\alpha$, to produce qualitatively different patterns of convergence, each reflecting a distinct balance between integration and compartmentalization. This allows the deliberation process to explore multiple regions of conceptual space (see Figure A2 in Appendix A). Aggregating the outputs across such heterogeneous runs expands the "epistemic range"—the number and distinctiveness of coherent solution clusters—available to the DM.

**Proposition 2**: Synthetic deliberation will outperform imagined deliberation when performance depends on dynamically balancing the compartmentalization and integration of perspectives during complex problem solving.

To illustrate tunability concretely, consider the green technology investment scenario adapted from, Koçak Puranam, & Yegin (2023b) that we introduce in Appendix B, where three executives hold conflicting views about an environmental investment. With low α (0.2), each executive maintains their distinct perspective—financial concerns, moral obligations, and technical doubts—exploring their reasoning deeply without interference. With high α (0.8), they quickly converge toward compromise positions. Most powerfully, synthetic deliberation can dynamically adjust α during deliberation, starting low to preserve diverse exploration then increasing it to promote integration, or run multiple deliberations with different α values to explore qualitatively different solution paths. This controlled modulation of perspective integration—impossible to achieve reliably through mental simulation—reveals different regions of the solution space that would otherwise remain hidden.

## 4.3. Problem Complexity: Ruggedness as a Boundary Condition for the Benefits of Externalization and Tunability

The advantages of synthetic deliberation over imagined deliberation should become most pronounced in rugged problem landscapes—solution spaces characterized by many local optima with low correlations in local fitness, arising from interactions among decision variables (Kauffman, 1993; Levinthal, 1997). In such landscapes, nearby solutions may differ greatly in quality, and local search tends to lead decision makers toward proximate peaks rather than toward the global optimum. The structure of many multi-attribute problems takes this form (Rittel and Weber, 1973). Because human reasoning often proceeds by analogy or incremental adjustment, individuals engaged in imagined deliberation are especially vulnerable to premature convergence when the solution space is large or poorly understood. Even deliberate attempts at compartmentalizing perspectives in imagined deliberation are undermined by cognitive constraints on working memory, attention, and self-distancing (Cowan, 2001; Nickerson, 1998).

In rugged landscapes, externalizability enables synthetic deliberation to preserve multiple, distinct search trajectories without interference, ensuring that exploration is not constrained by a single set of path-dependent mental moves. At the same time, tunability allows synthetic deliberation to control the timing and extent of integration across these diverse trajectories. This combination increases the likelihood of discovering novel, high-value solutions that lie between local peaks—solutions that imagined deliberation is less likely to generate due to early convergence or limited coverage of the solution space.

**Proposition 3**: The advantages of synthetic deliberation over imagined deliberation arising from externalizability and tunability will increase as the ruggedness of the decision maker's payoff landscape increases.

To illustrate ruggedness concretely, consider how interdependencies shape problem complexity in the green technology investment context from Appendix B. In a low ruggedness version of this problem, the financial, environmental, and technical dimensions would be largely independent—improving the technology's reliability would not dramatically affect its environmental benefits, and securing better financing terms would not alter technical requirements. Each executive could optimize their dimension separately with predictable results. However, the actual green technology investment represents a high ruggedness problem with complex interdependencies: improving technical reliability might require materials that reduce environmental benefits; securing government environmental subsidies could trigger regulatory requirements that affect both costs and technical specifications; and rushing implementation to meet environmental targets might compromise technical testing, cascading into higher maintenance costs that destroy financial viability. These interdependencies create multiple local peaks where synthetic deliberation's ability to maintain diverse search trajectories becomes essential.

## 5. GENERAL DISCUSSION

Synthetic deliberation uniquely advances cognitive flexibility by externalizing the dual processes that underlie cognitive flexibility. Unlike imagined deliberation, which is confined to an individual's mind, synthetic deliberation creates vivid, interactive representations of diverse viewpoints where compartmentalization and integration can be explicitly tuned. This technological scaffold allows for parallel processing of conflicting perspectives without the cognitive degradation and interference that typically occurs when humans attempt to maintain and work with multiple mental models simultaneously. Building on the green technology example discussed in our propositions, we provide a full practical demonstration of this process in Appendix B, using a customized chatbot built on GPT-based technology to simulate deliberation under varying integration parameters.

### 5.1. Synthetic Deliberation vs. Other Related Approaches

To fully appreciate the unique contributions of synthetic deliberation, it is important to contrast and differentiate it from related frameworks, such as digital twins, Agent-based models (ABMs), and AI-based devil's advocates.

*5.1.1. Synthetic Deliberation vs Digital Twins.* While both synthetic deliberation and digital twins utilize digital technology to enhance decision-making, their focus and mechanisms differ significantly. Digital twins aim to create a dynamic, two-way representation of real-world entities—such as products, processes, or organizations (Lyytinen, Weber, Becker, & Pentland, 2023). Their primary value lies in mirroring and predicting the behavior of these real-world counterparts, facilitating real-time monitoring, analysis, and intervention. Synthetic deliberation, by contrast, focuses on simulating deliberative discourse among agents representing diverse viewpoints on a problem. Leveraging LLMs, it creates a "synthetic" environment for exploring arguments, counter-arguments, and potential outcomes of different decision options. Unlike digital twins, which strive for a faithful representation of reality, synthetic deliberation adopts a more abstract and hypothetical approach, prioritizing the exploration of alternative perspectives and the mitigation of cognitive biases.

*5.1.2. Synthetic Deliberation vs Agent-Based Models.* Agent-based models (ABMs) employ Monte Carlo methods to generate probability distributions of potential outcomes by simulating interactions based on simple rules. The primary focus of ABMs is to explain complex outcomes through the interactions and dynamics arising from these rules (Knudsen, Levinthal, & Puranam, 2019). They are often used to model and predict the behavior of systems with many interacting agents, such as organizations, financial markets, ecosystems, or production lines.

In contrast, synthetic deliberation leverages LLM models to create environments that simulate deliberation, discourse, and dialogue among agents with different perspectives and interests. Rather than explaining specific outcomes or replicating real-world scenarios, synthetic deliberation aims to enhance human cognitive flexibility by simulating deliberative dialogue. It offers a structured framework for maintaining and integrating divergent perspectives, enabling richer explorations of alternative viewpoints. In essence, ABMs act as mirrors that reflect and, in the case of digital twins, potentially control reality, while synthetic deliberation serves as a platform for constructing and challenging potential realities through simulated discourse.

*5.1.3. Synthetic Deliberation vs AI-Based Devil's Advocate.* Recent research

demonstrates that groups assisted by LLM-based "devil's advocates" achieve higher accuracy in decision-making tasks, particularly when interactive AI tools are employed (Chiang, Lu, Li, & Yin, 2024). Both synthetic deliberation and the AI-powered devil's advocate utilize AI to enhance decision-making by introducing diverse perspectives, but they differ in scope and implementation. The AI-powered devil's advocate specifically focuses on improving group decision-making in contexts where AI already provides recommendations. Its primary goal is to prevent over-reliance on AI by prompting human group members to evaluate the AI's suggestions critically. This approach is grounded in a specific decision context and aims to optimize the interaction between human groups and AI systems.

In contrast, synthetic deliberation adopts a broader perspective. It seeks to enhance individual cognitive flexibility in addressing complex problems by simulating multiagent deliberation that goes beyond evaluating AI recommendations. This simulation exposes DMs to diverse viewpoints, arguments, and counter-arguments, fostering a more comprehensive understanding of the problem. Unlike the AI devil's advocate, which centers on group dynamics, synthetic deliberation aims to augment human mental simulation, which is often constrained by cognitive biases and limited working memory. By externalizing this internal process, synthetic deliberation provides a more robust, AI-assisted framework for exploring complex problems.

### 5.2. Theoretical and Practical Implications

Our dual-process framework uniquely addresses the dynamic and iterative temporal dimension of cognitive flexibility by explicitly combining compartmentalization and integration. Building upon Laureiro-Martínez and Brusoni's (2018) conceptualization of matching cognitive processes to problem types, we extend this foundation by introducing a dynamic perspective that captures how these processes operate iteratively over time. While existing models effectively describe cognitive flexibility at a point in time, our framework addresses the temporal dynamics of how perspectives evolve, separate, and reconcile, particularly in contexts characterized by tension or uncertainty.

Our model's emphasis on compartmentalization and integration also aligns with fundamental cognitive mechanisms identified in neuroscience research. For instance, Sigman and Dehaene (2008) demonstrated that the human brain employs both serial and parallel processing during complex tasks, with certain cognitive networks operating sequentially while others function simultaneously. This biological foundation supports our theoretical framework where compartmentalization (parallel processing of different perspectives) and integration (serial processing for synthesis) can coexist and complement each other. Furthermore, structured approaches like De Bono's (2017) six thinking hats method demonstrate how compartmentalization and integration can be systematically implemented in practice, allowing individuals to deliberately separate different modes of thinking before synthesizing insights into comprehensive solutions.

Our theory formalizes the logic explaining a crucial finding from Chi et al.'s (2017) research: observers of dialogue often demonstrate superior integration abilities compared to direct participants. These observers successfully merge multiple compartmentalized simulations into broader, cohesive meta-simulations. This effectiveness arises from dialogue's ability to promote both compartmentalization (by clearly delineating perspectives) and integration (by synthesizing conflicts and resolving viewpoints). Synthetic deliberation amplifies this observer advantage by creating a structured environment where multiple perspectives are clearly articulated and distinguished, yet available for integration. By positioning the decision maker as an observer of this process, we leverage the cognitive advantages identified in vicarious learning research while overcoming the practical limitations of arranging live dialogues for observation.

Our findings also contribute to the theoretical understanding of group decision-making and the role of AI-assisted processes. For example, Chiang et al. (2024) demonstrated in a randomized human-subject experiment that interactive LLM-powered devil's advocates are perceived as more collaborative and of higher quality. Similarly, Du, Li, Torralba, Tenenbaum, and Mordatch (2023) showed that structured debates between multiple LLMs can enhance model performance. Google's Notebook LM further illustrates the practical value of dialogue-based communication in AI systems, enabling the surfacing and debating of multiple perspectives through structured discourse. Notably, the interactive devil's advocate is particularly effective because it facilitates both the separation and synthesis of competing perspectives, aligning closely with our model of compartmentalization and integration. By dynamically engaging with group members and challenging AI recommendations, interactive advocates foster a structured deliberative process that supports unbiased exploration (compartmentalization) while enabling the reconciliation of diverse viewpoints (integration).

Our model extends beyond these findings by providing a broader theoretical framework for cognitive flexibility, which balances divergent and convergent thinking across diverse decision-making contexts. This generalizability makes our model a valuable tool not only for enhancing AI-assisted group decision-making but also for informing structured deliberation and synthesis in other complex problem-solving domains, advancing both theory and practice.

These theoretical insights have significant practical implications across multiple domains. For instance, in strategic business planning, synthetic deliberation can enable an organization's leaders to model interactions between different viewpoints for more comprehensive strategic responses. Policymakers can navigate competing stakeholder objectives in complex domains like climate policy by simulating impacts across different groups. In conflict resolution, it can alleviate cognitive and emotional barriers by simulating negotiations that lead to balanced solutions satisfying diverse interests.

While we have posed a contrast between synthetic and imagined deliberation, it is possible for the former to improve the latter if used in conjunction. Synthetic deliberation can amplify the benefits of mental simulation by facilitating engagement with diverse perspectives. Through interactions with LLM-powered agents, individual DMs can observe where different viewpoints diverge and how they interact, reason, and respond to challenges. This process, combined with incentives (Epley, Keysar, Van Boven, & Gilovich, 2004) and accountability (Tetlock, Skitka, & Boettger, 1989), potentially enables a richer, more vivid understanding of diverse perspectives than mental simulation alone. In multi-actor situations, an appreciation of the complexity of problems can enhance empathy toward other stakeholders (Galinsky & Moskowitz, 2000),

In addition, human decision-makers may be able to improve their own capability at internal deliberation. Synthetic deliberation provides a cognitively externalized model of compartmentalization and integration: simulated agents are explicitly bounded by distinct roles or priors, and their interactions are orchestrated such that their separation is both preserved and observable. Repeated exposure to these structured exchanges may allow human decision-makers to internalize the mechanics of maintaining viewpoint separability and careful integration. This vicarious learning can occur through processes of metacognitive mirroring and attentional cueing, as users observe how perspectives can be stabilized and compared without collapse. As a result, individuals who engage regularly with synthetic deliberation may develop enhanced capacity to replicate this compartmentalization internally, for instance, when later facing novel, multi-stakeholder problems where synthetic scaffolding is unavailable.

## 5.3. Limitations and Future Directions

While synthetic deliberation offers promising insights, it faces several limitations. The AI-driven agents may reflect biases in their training data, potentially skewing the representation of perspectives. Generated arguments might lack depth or authenticity, and complex perspectives could be oversimplified. The quality of synthetic deliberation thus unavoidably depends on the capabilities of the underlying AI models, with models producing less coherent outputs. Ethical concerns around privacy, autonomy, and manipulation must also be addressed (Safdar, Banja, & Meltzer, 2020). Addressing these limitations through model refinement, bias mitigation, transparency, and ethical safeguards is critical for improving reliability and applicability.

Beyond these technical limitations, the adoption of synthetic deliberation faces significant behavioral challenges in practice. A critical tension emerges in how users integrate the simulated perspectives. Simulation models can appear deceptively credible and convincing, leading to over-reliance where users uncritically accept simulated perspectives without maintaining their own viewpoint (Zhai, Wibowo, & Li, 2024). Synthetic deliberation carries the risk of excessive cognitive offloading—a tendency for decision makers to defer too much of the reasoning process to the system. Research on AI-assisted writing tasks shows that when users rely heavily on LLMs without first engaging cognitively with the problem, they exhibit lower semantic encoding, weaker memory consolidation, and reduced executive self-monitoring, often copying outputs verbatim rather than integrating them with their own reasoning (Kosmyna et al, 2025). Related findings in educational contexts suggest that while LLM use reduces cognitive load, it can also diminish the depth and diversity of reasoning if users fail to evaluate the AI's output critically (Stadler et al, 2024). This disengagement can lead to skill atrophy, where the human's own capacity for compartmentalization and integration decays over time.

  While we acknowledge this risk, the very nature of synthetic deliberation is such that it is likely to sustain user engagement: observing an active, evolving debate between synthetic agents is inherently stimulating and can invite curiosity, scrutiny, and reflection rather than passive acceptance. If designed to require active stance-taking, justification, and critical comparison between agent outputs, synthetic deliberation can not only scaffold immediate decision quality but also cultivate the very cognitive flexibility it aims to enhance. The challenge—and opportunity—lies in configuring these systems so that they augment, rather than replace, the uniquely human capacities for judgment, synthesis, and creative problem-solving.

innovation. *Strategic Management Journal*, *36*(10), 1435-1457.

Karim, S., & Kaul, A. (2015). Structural recombination and innovation: Unlocking intraorganizational knowledge synergy through structural change. *Organization Science*, *26*(2), 439-455.

Katila, R., & Ahuja, G. (2002). Something old, something new: A longitudinal study of search behavior and new product introduction. *Academy of Management Journal*, *45*(6), 1183-1194.

Kauffman, S. A. (1993). *The Origins of Order: Self-Organization and Selection in Evolution*. Oxford University Press.

Keeney, R. L., & Raiffa, H. (1993). *Decisions with multiple objectives: Preferences and value tradeoffs (2nd ed.)*. Cambridge, England: Cambridge University Press.

Kirkpatrick, S., Gelatt Jr, C. D., & Vecchi, M. P. (1983). Optimization by simulated annealing. *Science*, 220(4598), 671-680.

Knudsen, T., A. Levinthal, D., & Puranam, P. (2019). A model is a model. *Strategy Science*, *4*(1), 1-3.

Koçak, Ö., Levinthal, D. A., & Puranam, P. (2023a). The dual challenge of search and coordination for organizational adaptation: How structures of influence matter. *Organization Science*, *34*(2), 851-869.

Koçak, Ö., Puranam, P., & Yegin, A. (2023b). Decoding cultural conflicts. *Frontiers in Psychology*, *14*, 1166023.

Kogut, B., & Zander, U. (1993). Knowledge of the firm and the evolutionary theory of the multinational corporation. *Journal of International Business Studies*, *24*, 625-645.

Kotovsky, K., Hayes, J. R., & Simon, H. A. (1985). Why are some problems hard? Evidence from Tower of Hanoi. *Cognitive psychology*, 17(2), 248-294.

Kotovsky, K., & Simon, H. A. (1990). What makes some problems really hard: Explorations in the problem space of difficulty. *Cognitive psychology*, 22(2), 143-183.

Kozlowski, A. C., & Evans, J. (2025). Simulating Subjects: The Promise and Peril of Artificial Intelligence Stand-Ins for Social Agents and Interactions. *Sociological Methods & Research*, 00491241251337316.

Krems, J. F. (2014). Cognitive flexibility and complex problem solving. In: *Complex Problem Solving* (pp. 201-218). Psychology Press.
31

the usefulness of inventions and knowledge-base malleability. *Administrative Science Quarterly*, *53*(2), 333-362.

Zhai, C., Wibowo, S., & Li, L. D. (2024). The effects of over-reliance on AI dialogue systems on students' cognitive abilities: a systematic review. *Smart Learning Environments*, *11*(1), 28.

**Appendix A: Simulations of the model of the dual processes underlying cognitive flexibility**

This appendix provides a computational model for the deliberation process described in Section 3, and we demonstrate the tunability advantage of synthetic deliberation in Section 4.2. The code to reproduce the key results is available online.

**A.1 Model Structure**

To computationally implement our formal model, we adopt the NK landscape framework (see Kauffman, 1993; Levinthal, 1997), which has been canonical for complex problem-solving (e.g., Fleming & Sorenson, 2004; Katila & Ahuja, 2002; Afuah & Tucci, 2012; see Boussioux et al., 2024 for a recent study). In this framework, each solution (or each position in the landscape) is represented by $N$ binary components, and the payoff for each component is contingent upon $K$ other components. When $K = 0$, each component is independent of the others in determining its payoff, and the fitness landscape is featured as a single-peaked landscape in which agents with local search can discover the global peak. As $K$ increases, the number of peaks in the landscape increases, and thus local search may entrap agents into suboptimal solutions. With this landscape, we (1) randomly allocate $m$ agents at the initial period, (2) have them conduct local search in parallel until they reach local peaks, (3) randomly choose one agent to share its proposal, and (4) let others incorporate it with the probability of $\alpha$ for each component. The processes (2) ~ (4) are repeated for $T$ periods.

**A.2 Key Results**

Based on the above model, we examine the number of solutions discovered for varying levels of $\alpha$. For model parameters, we use $N = 10$, $K \in \{0, 1, 5, 9\}$, $T = 1{,}000$, and $m = 5$. All repeated simulations involve 10,000 runs.

In Figure A1, we demonstrate the impact of the temporal adjustment of the integration parameter by comparing two cases: one with a constant $\alpha = 0.5$ and the other with $\alpha$ linearly increasing from 0 to 1. While the level of the integration parameter is on average 0.5 in both cases, our result shows that synthetic deliberation with the increasing $\alpha$ produces more solutions than the constant $\alpha$, proving the benefits arising from the within-run tunability.

Figure A2 describes the relationship between the integration parameter ($\alpha$) and the number of solutions discovered. In Figure A2a, we assume that agents have aligned incentives, meaning that they agree on the value of the proposed alternative while exploring different areas of the solution space. In Figure A2b, we relax this assumption by allowing them to have diverging evaluations for a given alternative due to heterogeneous value systems.

We first confirm that the variety of solutions discovered has an inverted U-shaped relationship with $\alpha$ in both cases. Second, our result shows that the benefits of cognitive flexibility—balancing compartmentalization and integration—are contingent upon the degree of complexity ($K$) (Proposition 3). When $K$ is low, the fitness landscape is characterized as a simple and single-peaked landscape in which the global optimum is reachable for a single agent with local search. In contrast, as $K$ increases, individual agents may be entrapped at local peaks, and thus cognitive flexibility is needed to balance parallel exploration and synthesis of diverse perspectives. Lastly, we find that the optimal integration level depends on the problem complexity and divergence in viewpoints. Thus, finding the optimal point requires repeated, independent experiments, offering the between-run tunability advantage to synthetic deliberation.

**Figure A1. The impact of the within-run tunability on the number of solutions discovered**

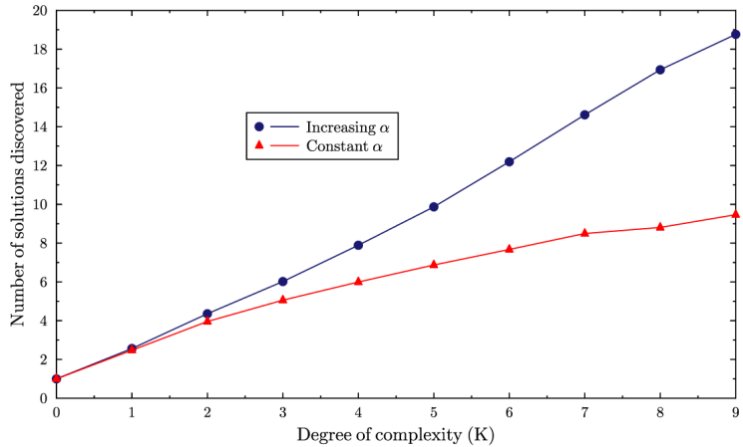

**Figure A2. The integration parameter ($\alpha$) and the number of solutions discovered**

(a) When agents have aligned incentives  (b) When agents have misaligned incentives

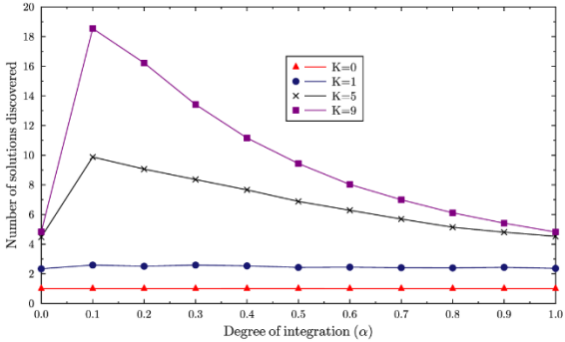 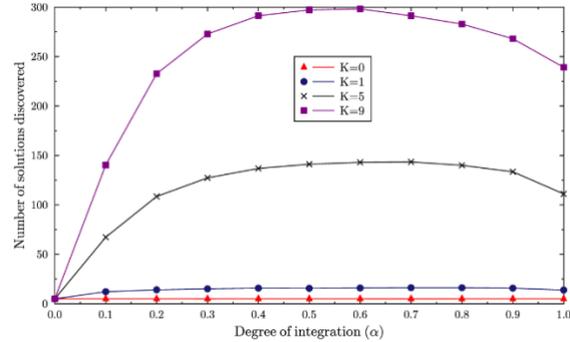

**Appendix B: Demonstration of Synthetic Deliberation**

This appendix offers a practical demonstration using a customized bot built on GPT4o-based technology (https://chatgpt.com/share/6776864e-db20-8008-b555-d2ca8a1009bc). The user prompt is provided below (Appendix A1), and the sample output is available on the Open Science Framework (https://osf.io/sk45r/?view_only=8af3c825e5d3430eab9e2c9351e244d6).

**B.1 User Prompt**

You will simulate a deliberation between three executives discussing a green technology investment, based on their independently formed views as described below. Use this vignette as their starting positions:

[Vignette]
In thinking independently and privately about the proposed green technology investment, A, B, and C each had distinctive concerns and priorities. A acknowledged that the project would substantially reduce carbon emissions well below legal requirements but maintained that such reductions shouldn't come at the expense of shareholder value, noting that returns would be significantly lower compared to alternative investments. B, while agreeing about the emission reductions, argued passionately that the company had a moral obligation to minimize environmental impact regardless of legal minimums, viewing the investment as essential despite its lower financial returns. C, however, questioned the fundamental premise – while supporting environmental initiatives in principle, they had serious doubts about the technology's effectiveness, citing inconsistent test results and a lack of large-scale implementation data. Their technical analysis suggested the claimed emission reductions were likely overstated due to maintenance issues and operational constraints not accounted for in the initial projections.

Your task is to simulate a synthetic deliberation between these three executives, with their willingness to adjust positions, alpha ($\alpha$), set as [input value between 0 and 1, inclusive].

**B.2 Sample Output**

The sample output of this prompt for alpha=0, 0.5, and 1 is available on the Open Science Framework: https://osf.io/sk45r/?view_only=8af3c825e5d3430eab9e2c9351e244d6